%% file: HRLASC.tex
\documentclass[sigconf]{acmart}
\AtBeginDocument{%
  }
\usepackage{amsmath}
\usepackage{amsfonts}
\usepackage{placeins}
\usepackage{tabularx}
\usepackage{booktabs}
\usepackage[linesnumbered,ruled,vlined]{algorithm2e}
\setcopyright{cc}
\setcctype{by}
\copyrightyear{2026}
\acmYear{2026}
\acmConference[WWW '26]{Proceedings of the ACM Web Conference 2026}{April 13--17, 2026}{Dubai, United Arab Emirates}
\acmBooktitle{Proceedings of the ACM Web Conference 2026 (WWW '26), April 13--17, 2026, Dubai, United Arab Emirates}
\acmISBN{979-8-4007-2307-0/2026/04}
\usepackage{graphicx}
\usepackage{caption}
\usepackage{subcaption}
\usepackage[english]{babel}
\usepackage{amsthm}
\usepackage{listings}
\usepackage{xcolor}
\acmSubmissionID{507}
\begin{document}

\title{DARA: Few-shot Budget Allocation in Online Advertising via In-Context Decision Making with RL-Finetuned LLMs}


\author{Mingxuan Song}
\affiliation{
    \institution{Peking University}
    \institution{School of Computer Science}
    \city{Beijing}
    \country{China}
}
\email{songmingxuan@stu.pku.edu.cn}

\author{Yusen Huo}
\affiliation{
    \institution{Alibaba Group}
    \city{Beijing}
    \country{China}
}
\email{huoyusen.huoyusen@alibaba-inc.com}

\author{Bohan Zhou}
\affiliation{
    \institution{Peking University}
    \institution{School of Computer Science}
    \city{Beijing}
    \country{China}
}
\email{zhoubh@stu.pku.edu.cn}

\author{Shenglin Yin}
\affiliation{
    \institution{Peking University}
    \institution{School of Computer Science}
    \city{Beijing}
    \country{China}
}
\email{yinsl@stu.pku.edu.cn}

\author{Zhen Xiao}
\authornote{Corresponding author.}
\affiliation{
    \institution{Peking University}
    \institution{School of Computer Science}
    \city{Beijing}
    \country{China}
}
\email{xiaozhen@pku.edu.cn}

\author{Jieyi Long}
\affiliation{
    \institution{Theta Labs, Inc.}
    \city{San Jose}
    \country{USA}
}
\email{jieyi@thetalabs.org}

\author{Zhilin Zhang}
\authornotemark[1]
\affiliation{
    \institution{Alibaba Group}
    \city{Beijing}
    \country{China}
}
\email{zhangzhilin.pt@alibaba-inc.com}

\author{Chuan Yu}
\affiliation{
    \institution{Alibaba Group}
    \city{Beijing}
    \country{China}
}
\email{yuchuan.yc@alibaba-inc.com}

\renewcommand{\shortauthors}{Mingxuan Song, et al.}


\begin{abstract}
Optimizing the advertiser’s cumulative value of winning impressions under budget constraints poses a complex challenge in online advertising, under the paradigm of AI-Generated Bidding (AIGB). Advertisers often have personalized objectives but limited historical interaction data, resulting in few-shot scenarios where traditional reinforcement learning (RL) methods struggle to perform effectively. Large Language Models (LLMs) offer a promising alternative for AIGB by leveraging their in-context learning capabilities to generalize from limited data. However, they lack the numerical precision required for fine-grained optimization. To address this limitation, we introduce \textbf{GRPO-Adaptive}, an efficient LLM post-training strategy that enhances both reasoning and numerical precision by dynamically updating the reference policy during training. Built upon this foundation, we further propose \textbf{DARA}, a novel dual-phase framework that decomposes the decision-making process into two stages: a few-shot reasoner that generates initial plans via in-context prompting, and a fine-grained optimizer that refines these plans using feedback-driven reasoning. This separation allows DARA to combine LLMs’ in-context learning strengths with precise adaptability required by AIGB tasks. Extensive experiments on both real-world and synthetic data environments demonstrate that our approach consistently outperforms existing baselines in terms of cumulative advertiser value under budget constraints.
\end{abstract}


% \ccsdesc[500]{Theory of computation~Computational pricing and auctions}



\maketitle

\input{chaps/introduction}
\input{chaps/backgroundrelatedwork}
\input{chaps/RLAMChain}
\input{chaps/RLAM}
\input{chaps/evaluation}

\section{Conclusion}

In this work, we explore the challenging task of few-shot budget allocation in auction advertising, where data scarcity presents significant obstacles to effective decision making. Our solution features a dual-phase LLM architecture that separates few-shot reasoning from numerical optimization. To further enhance the decision quality of both agents, we introduce GRPO-Adaptive, an improved RL fine-tuning algorithm that dynamically updates the KL regularization anchor to stabilize and guide policy learning. Through comprehensive experiments across real-world and simulated environments, we demonstrate that our method significantly outperforms strong baselines. Overall, our work offers a promising direction for combining few-shot LLM reasoning with RL fine-tuning in budget allocation problems.

\begin{acks}
The authors would like to thank the anonymous reviewers for their comments. This work was supported by the National Key R\&D Program of China under Grant 2023YFB2703800, and supported by Alibaba Group through Alibaba Innovative Research Program. The contact author is Zhen Xiao and Zhilin Zhang.
\end{acks}

\bibliographystyle{ACM-Reference-Format}
\balance
\bibliography{sample-base}

\appendix
\input{chaps/appendix/SafetyAnalysis}

\end{document}

%% file: chaps/introduction.tex
\section{Introduction}
\label{sec:intro}
Maximizing the cumulative value of winning impressions is a crucial sequential decision-making challenge in online advertising, where the system must automatically generate bidding decisions on behalf of advertisers to optimize long-term performance~\cite{nuara2022online,su2024auctionnet}. Under the emerging paradigm of AI-Generated Bidding (AIGB), A growing line of work has advocated for decomposing bidding strategies into budget allocation and bid decision, treating the former as a global planning task across time periods~\cite{guo2024generative,hajiaghayi2022analysis,duan2025adaptable}. This separation allows the system to first produce high-level budget plans and then optimize bid decisions within the budget constraints, well suited for this task~\cite{avadhanula2021stochastic}. In budget allocation task, advertisers must determine how to allocate limited budgets across multiple time periods to maximize return on investment (ROI), while dynamically adapting to temporal cost-reward patterns from shifting user behavior and dynamics of online environments~\cite {liu2020effective}. 

This problem exhibits several key characteristics—sequential dependency, delayed rewards, and non-stationary environment dynamics—that make Reinforcement Learning (RL) particularly well suited~\cite{li2018efficient}. RL methods, including Q-learning~\cite{li2018efficient,li2024trajectory} and hierarchical RL~\cite{wang2023hibid,duan2025adaptable}, have been introduced to address these challenges by learning optimal allocation strategies from historical interactions. In advertising applications, advertisers often have personalized goals and dynamic advertiser preferences, which naturally give rise to few-shot or cold-start scenarios where only limited data is available for each task~\cite{soboleva2025optimizing,guo2024generative}. However, RL‑based methods typically require extensive environment interactions and large volumes of task-specific data to train. Therefore, the resulting policies often fail to generalize well and cannot adapt quickly when faced with new or dynamically changing environments~\cite{nguyen2020deep,padakandla2021survey,dong2022simple}.

Recent advances in large language models (LLMs) have opened new possibilities for decision making in low-data regimes~\cite{lin2023layoutprompter,zhang2023making}. The in-context learning of LLMs is to perform new tasks by conditioning on a few annotated examples presented in the input prompt. This ability allows LLMs to rapidly adapt to new tasks with limited data, making them especially suitable for personalized budget planning in data-scarce AIGB environments~\cite{seedat2023curated}. However, despite their flexibility in few-shot prompting, LLMs often exhibit insufficient numerical sensitivity, leading to a lack of the fine-grained precision necessary for structured optimization tasks such as budget allocation~\cite{hadillms,klouda2025meta}. Furthermore, our ablation studies in Section~\ref{sec:ablation} reveal that the in-context learning capability of LLMs is limited when applied to complex decision-making tasks: a single-stage prompting setup struggles to both capture underlying few-shot data regularities and execute fine-grained budget optimization. These limitations underscore the need for a new framework that combines the generalization strength of LLMs with the numerical precision and adaptability of reinforcement learning (RL), thereby enabling more robust and effective budget allocation under realistic constraints. 

To bridge this gap, we revisit the structure of the budget allocation task and observe that it naturally decomposes into two sub-problems with distinct characteristics. The first phase involves generalizing from limited historical patterns to derive high-level allocation intentions, while the second phase requires fine-grained adjustments based on dynamic environment feedback. Existing approaches either treat the task as a single phase optimization problem or rely on monolithic RL training, failing to capture this structural separation~\cite{duan2025adaptable,song2022domain,song2025aero}. We posit that an effective solution should explicitly separate the generalization and optimization processes, aligning model capabilities with the distinct demands of each phase. In addition, RL can be employed to fine-tune the LLM, further enhancing its reasoning capacity and optimization precision across phases~\cite{kaufmann2024survey,li2024spring}. To overcome the limited amount of few-shot data, we draw inspiration from the idea of dynamics randomization in sim-to-real learning~\cite{peng2018sim}, and design a simulation environment tailored for budget allocation. This environment is modeled after real-world data distributions and is capable of continuously generating diverse allocation scenarios. % Through this randomized training, the model learns a robust and generalizable policy.

Based on the above analysis, we propose DARA, a novel \textbf{D}ual-phase \textbf{A}daptive \textbf{R}easoning and \textbf{A}llocation framework for few-shot budget allocation in online advertising under the AIGB paradigm. We decompose the task into two phases and accordingly design a dual-agent architecture: a \textbf{Few-shot Reasoner} that generates initial budget plans based on few-shot in-context data, and a \textbf{Fine-grained Optimizer} that incrementally refines the plan through feedback-driven reasoning. This design follows the principle that early-phase decisions benefit from few-shot generalization, while later-phase refinements require numerically sensitive, feedback-aware adjustments. 

Moreover, purely prompt-based reasoning in LLMs often lacks numerical sensitivity and fine-grained control~\cite{loya2023exploring}. To overcome this limitation, we introduce a reinforcement learning fine-tuning strategy to enhance the decision quality of both LLMs in our framework. We propose GRPO-Adaptive, which periodically updates the reference model during training. By comparing the current policy against a more recent baseline rather than a fixed one, GRPO-Adaptive enables more flexible and effective policy improvements.

We also integrate a dual-environment setup into our method: (1) a real-world environment derived from enterprise-scale advertising data, and (2) a simulation environment designed based on controllable polynomial functions. The simulation environment enables the continuous generation of diverse allocation scenarios, enhancing policy robustness through exposure to varied conditions. Experiments show that DARA consistently outperforms baselines. 

In summary, our contributions are as follows.
\begin{itemize}
\item We introduce DARA, a dual-phase LLM-based architecture for budget allocation, enabling robust and interpretable decision making under few-shot conditions.
\item We propose GRPO-Adaptive, a novel RL algorithm that dynamically updates the reference model to enhance the reasoning ability of LLM.
\item We design a simulation environment that mimics real-world allocation dynamics and continuously generates diverse budget scenarios. This simulation enables the model to learn robust and generalizable policies from limited data.
\end{itemize}

%% file: chaps/backgroundrelatedwork.tex
\section{Background and Related Work}
\label{sec:related}

\subsection{Preliminaries}
% This objective is typically measured by the Return On Investment (ROI), defined as the ratio between cumulative value and total cost. In online advertising, the total budget $B$ must be distributed across $T$ discrete time periods. 
In Real Time Bidding (RTB), advertisers aim to maximize the cumulative value of winning impressions within a fixed budget $B$. Let $b_t$ denote the budget allocated to period $t$, and $v_t(b_t)$ denote the expected return when allocating a budget of $b_t$ in time period $t$. Then, the ROI is:
\begin{equation}
\text{ROI} = \frac{\sum_{t=1}^T v_t(b_t)}{\sum_{t=1}^T b_t} = \frac{\sum_{t=1}^T v_t(b_t)}{B}
\end{equation}
Since $B$ is fixed, maximizing ROI is equivalent to maximizing the total return $\sum_{t=1}^T v_t(b_t)$ under the constraint $\sum_{t=1}^T b_t = B$.

\vspace{0.5em}
\textbf{Assumption.}  
Each function $v_t(b_t)$ is assumed to be differentiable, strictly increasing, and concave in $b_t$. This implies that the marginal ROI, given by $v_t'(b_t)$, is positive and strictly decreasing with respect to budget—capturing the realistic effect of diminishing returns.

\vspace{0.5em}
\textbf{Optimality Condition.}  
Under the above assumptions, the optimal allocation satisfies that the marginal ROI is equal across all time periods:
\begin{equation}
    v_1(b_1^*) = v_2(b_2^*) = \cdots = v_T(b_T^*)
\end{equation}
where $b_t^*$ denotes the optimal budget for period $t$.

\vspace{0.5em}
\textbf{Proof Sketch.}  
Suppose, for contradiction, that there exist two periods $i$ and $j$ such that $v_i'(b_i) > v_j'(b_j)$. By shifting a small amount of budget $\delta > 0$ from period $j$ to period $i$, the increase in total return is:
\begin{equation}
v_i(b_i + \delta) - v_i(b_i) > v_j(b_j) - v_j(b_j - \delta)
\end{equation}
due to $v_i'(b_i) > v_j'(b_j)$ and concavity. This strictly increases the total return under the same budget, contradicting optimality. Therefore, all marginal ROIs must be equal at optimality. The complete proof is provided in Appendix \ref{appendix:A}. Accordingly, maximizing expected return under a fixed budget can be approximated by minimizing the variance of marginal ROI across time periods.

\subsection{Related Work}
% Traditional methods face inherent limitations in modern advertising scenarios~\cite{ou2023survey}. Rule-based heuristics, although interpretable and easy to implement, often fail to capture the stochasticity and evolving patterns in user behavior and auction dynamics~\cite{jones2002combinatorial,mukherjee2017apriori}. 

% Intuitively, to maximize total ROI under a fixed budget, the optimal allocation should balance the marginal ROI across periods—i.e., additional budget spent in any time period should yield similar returns. Otherwise, shifting budget from a lower marginal ROI period to a higher one would lead to an immediate gain, implying suboptimality~\cite{su2024spending}. Therefore, equalizing marginal ROI is a necessary condition for optimal allocation~\cite{wang2022roi}.

Budget allocation has long been a core problem in auction-based advertising systems, where advertisers aim to distribute limited budgets to maximize their overall return on investment (ROI)~\cite{borgs2007dynamics,cao2025map}. 
Early approaches in this domain typically rely on hand-crafted heuristics or rule-based strategies, which often fail to generalize across dynamic environments and complex auction conditions~\cite{jones2002combinatorial,mukherjee2017apriori}. To address these limitations, reinforcement learning (RL) has been widely adopted in recent years to learn optimal allocation strategies through interactions with the environment~\cite{li2018efficient,duan2025adaptable,li2024trajectory,wang2023hibid}. For example, in Q-MCKP~\cite{li2018efficient}, the budget space is discretized into multiple bins, and Q-value functions are learned independently for each stage. A multi-choice knapsack problem is then solved to determine the final allocation. This stage-wise independence overlooks potential interactions between time periods, which may be critical for achieving long-term budget efficiency. HiBid~\cite{wang2023hibid} adopts a hierarchical RL framework where a high-level policy allocates budget across time periods, and a low-level bidding agent learns to bid effectively within each period under budget constraints. However, it lacks mechanisms for continuous online adaptation or rapid fine-tuning in dynamic environments. Similarly, ABPlanner~\cite{duan2025adaptable} employs an LSTM-based RL planner to allocate budget across time periods. As an RL approach, it struggles to perform well under limited data and shifting advertiser conditions.

Recent advances in large language models (LLMs) have shown promising capabilities in low-data regimes through in-context learning~\cite{brown2020language,li2025multi,li2025rethinking}. Conditioning on a few examples, LLMs can perform structured decision-making tasks without explicit gradient updates~\cite{reynolds2021prompt}. These properties have inspired the application of LLMs to planning, control, and economic optimization tasks, where generalization and reasoning are critical.
However, vanilla prompting of LLMs often lacks the numerical sensitivity required for precise budget allocation. To address this, DPO~\cite{rafailov2023direct,bai2024ormer} has been proposed. DPO aligns LLM behavior with human preferences by learning from pairwise comparisons between preferred and rejected responses, offering better generalization in under-constrained tasks. More recently, Group Relative Proximal Optimization (GRPO)~\cite{guo2025deepseek} extends this idea to structured decision-making settings by introducing group-wise KL regularization, enabling more stable and sample-efficient LLM fine-tuning under noisy or feedback-driven supervision.

%% file: chaps/RLAMChain.tex
\section{LLM-based Dual-phase Decision Making}
\label{sec:AEROChain}

\subsection{Environment Modeling}

\begin{figure}[h]
    \centering
    \begin{subfigure}[t]{0.35\textwidth} % First subfigure
        \centering
        \includegraphics[width=\textwidth]{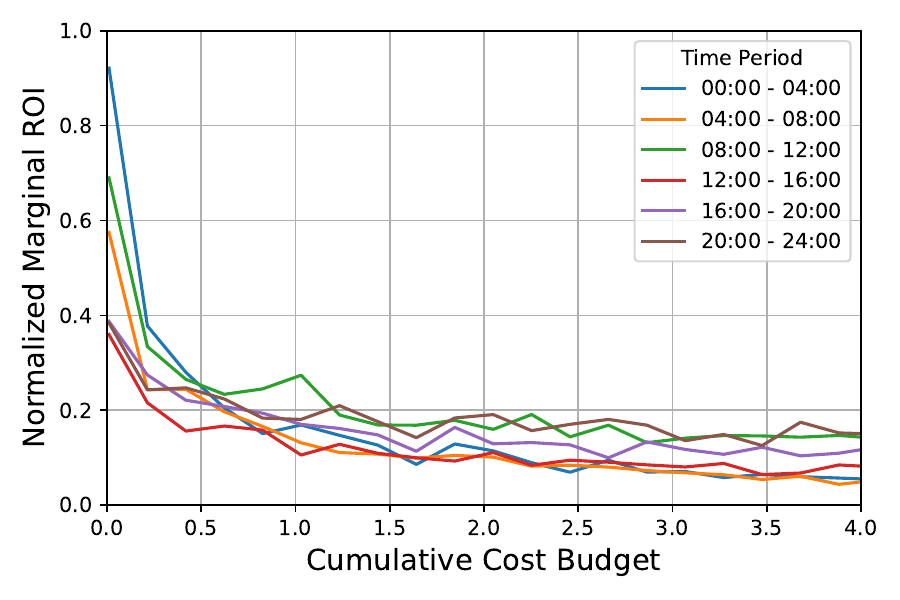}
        \caption{Real-World Data Environment.}
        \label{fig:real_env}
    \end{subfigure}
    % \hfill
    \begin{subfigure}[t]{0.35\textwidth} % Second subfigure
        \centering
        \includegraphics[width=\textwidth]{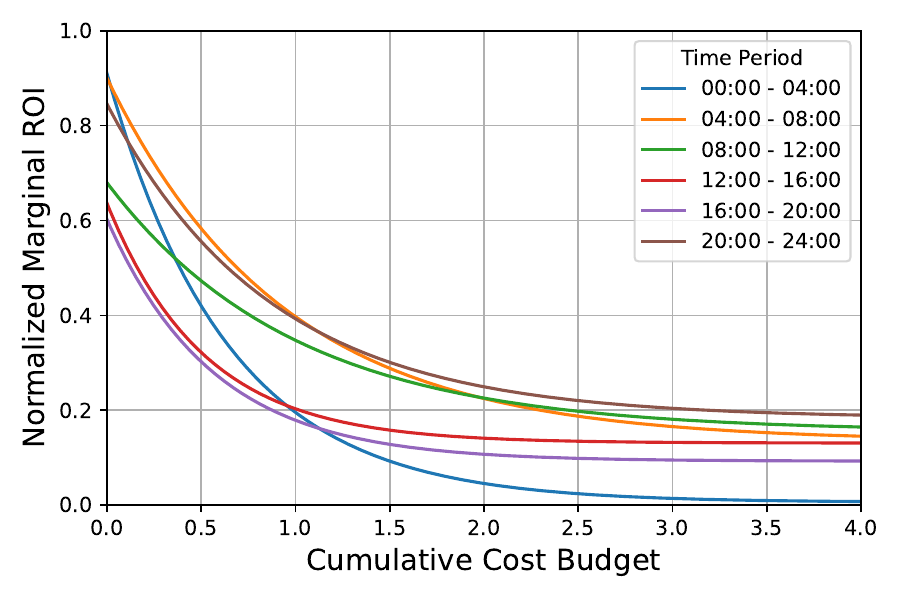}
        \caption{Synthetic Data Environment.}
        \vspace{-5pt}
        \label{fig:poly_env}
    \end{subfigure}
    \captionsetup{belowskip=-10pt}
    \caption{Data Environment.}
    \label{fig:ab_fig}
\end{figure}

To address the challenges of personalized budget planning, we incorporate a modeling strategy to support robust and generalizable policy learning under limited data, as illustrated in Figure~\ref{fig:ab_fig}. The rationale lies in the empirical statistics derived from real-world data: due to the cost-value curve of budget, the ROI slope typically decreases with increased budget. In addition, different time periods exhibit varying ROI patterns, making dynamic allocation essential. This prototype builds upon empirical patterns observed in online advertising data and provides two distinct environment modeling approaches: 

\begin{itemize}
  \item \textbf{Real-World Data Environment}: This environment is constructed directly from real-world online advertiser data, ensuring that environmental features, cost consumption behavior, and ROI dynamics align closely with actual scenarios. An example environment of the resulting marginal ROI curves is illustrated in Figure~\ref{fig:real_env}.
  
  \item \textbf{Synthetic Data Environment}: Given the limitations of real-world data, such as the inability to flexibly extend to budget strategies across varying time periods, we construct a simulation environment based on functional modeling to replicate the relationship between budget allocation and marginal ROI. This approach enables controlled experimentation and improves the generalizability of learned policies. As illustrated in Figure~\ref{fig:poly_env}, the marginal ROI function $MROI$ is designed to decrease monotonically for budget $b > 0$ until it reaches zero at a predefined budget:
\begin{equation}
{MROI_i}(b) = \max \left\{ F_i(b),\; 0 \right\}, \,\, 0 \leq b \leq B, \, i =1 \,\,to \,\, T,
\end{equation}
where $F_i(b)$ can be either a polynomial or an exponential function, $B$ is the total budget, and $T$ is the number of periods. This function is continuous and differentiable, making it well suited for optimization and analytical analysis.
\end{itemize}

These environmental modeling strategies provide two key advantages: 
(1) High fidelity to real-world settings: the cost and ROI patterns align closely with actual market behaviors, ensuring practical relevance. 
(2) Performance benchmarking: budget ceilings are manually defined to establish clear evaluation baselines, facilitating comparisons across models and policies. Additionally, the environment supports flexible adjustment of sequence lengths and environment parameters, enabling in-depth analysis of model behaviors under varying levels of complexity.

\subsection{Problem Formulation}

We formulate the few-shot budget allocation task as follows. Given a few-shot dataset $\mathcal{H}$ of $n$ previous episodes:

\begin{align}
\mathcal{H} &= \left\{ \left( \mathbf{b}^{(i)}, \mathbf{m}^{(i)} \right) \right\}_{i=1}^{n}, \\
\mathbf{b}^{(i)} &= \left( b^{(i)}_1, \dots, b^{(i)}_T \right), \\
\mathbf{m}^{(i)} &= \left( MROI^{(i)}_1, \dots, MROI^{(i)}_T \right)
\end{align}

\noindent where each $\mathbf{b}^{(i)}$ is a budget allocation vector across $T$ time periods and $\mathbf{m}^{(i)}$ is the corresponding marginal ROI vector, the goal is to generate a new allocation $\mathbf{b} = (b_1, \dots, b_T)$ that satisfies the total budget constraint:

\begin{equation}
\sum_{t=1}^{T} b_t = B, \quad b_t \geq 0,
\end{equation}

\noindent and minimizes the variance of the resulting $T$ $MROI$. This objective must be achieved with a minimal number of exploration trials, reflecting the few-shot constraint inherent in practical online advertising scenarios.

\subsection{Few-shot Prompting}

In advertising budget planning tasks, data is limited and deployment scenario changes dynamically. LLMs possess strong language understanding and generalization capabilities. They can complete complex tasks with only a few contextual examples, offering a promising alternative for handling small-sample budget allocation challenges. We propose a few-shot prompting framework that encodes historical episode data into prompts, allowing the model to generate effective strategies for current allocation tasks. 

To support task comprehension and improve response consistency, we design a structured prompt format that explicitly includes the following components: the task objective, few-shot data, trial records, and a clear output format specification. The objective section defines the goal of minimizing reward variance across periods while strictly adhering to a total budget constraint. The historical data block $\mathcal{H}$, which serves as few-shot data collected from advertisers, provides examples of previous budget allocation records and their associated ROI outcomes, helping the model learn from past patterns. The records block contains previously attempted strategies to guide policy refinement. The output specification section instructs the model to return a concrete allocation vector along with reasoning for the proposed strategy, ensuring interpretability and executability. Table~\ref{tab:fewshot-prompt} illustrates a concrete implementation example of our prompting approach. Two complete examples of such prompts are provided in the Appendix \ref{apen:PromptExamples}.

\begin{table}
\centering
\begin{tabular}{p{0.95\linewidth}}
\toprule
\ttfamily
You are assigned a budget allocation task consisting of \{NUM\} time periods. Each time period corresponds to a specific reward value. It is known that the relationship between budget and reward is approximately monotonically decreasing... \textless{}few-shot\textgreater{} \textless{}trial\_records\textgreater{} \textless{}objective\textgreater{} \textless{}output\_format\textgreater{} \\
\bottomrule
\end{tabular}
\caption{Template for Few-shot Prompting. \{NUM\} will be replaced with the specific number of periods.}
\label{tab:fewshot-prompt}
\end{table}

\subsection{Multi-Agent Paradigm}

In budget allocation tasks, we initially attempted to use a single LLM to perform direct decision-making for few-shot budget planning. However, we found that a single LLM struggles to simultaneously achieve rapid few-shot learning and precise policy modeling. This limitation was clearly validated in our ablation studies in Section \ref{sec:ablation}, which show that relying solely on a single LLM leads to performance ceilings in complex decision-making scenarios.
\begin{figure}
    \centering
    \includegraphics[width=0.48\textwidth]{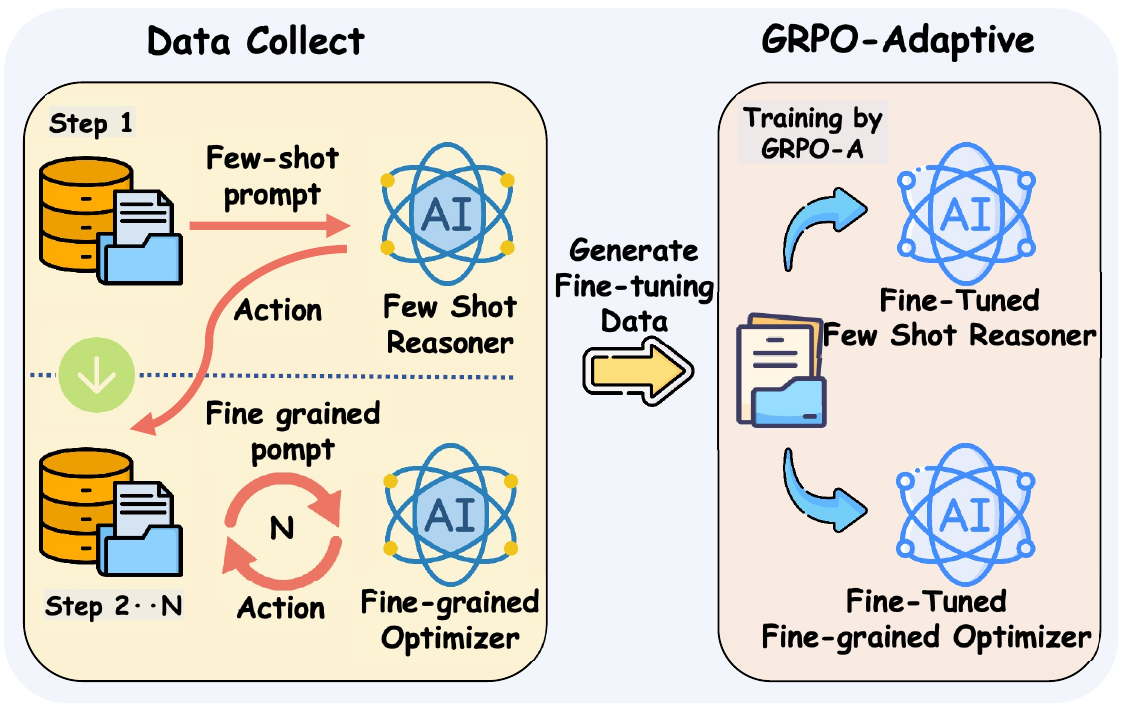}
    % \vspace{-2em}
    \captionsetup{belowskip=-5pt}
    \caption{The training workflow of DARA.}
    \label{fig:dataStructure}
    \end{figure}

To overcome the limitations of using a single LLM for end-to-end budget planning, we propose a Dual-Phase Cooperative Agent Architecture that decomposes the task along the temporal axis into two distinct phases: the first few-shot reasoning phase and subsequent decision-making phases, as illustrated in Figure~\ref{fig:dataStructure}. This design is motivated by the intuition that different stages of budget allocation require different reasoning capabilities—initial planning relies heavily on few-shot generalization, while later stages require fine-grained optimization based on early outcomes. Accordingly, we introduce two independent yet complementary LLM agents:

\textbf{Few Shot Reasoner} is responsible for the first episode budget allocation based on a few-shot prompt. It utilizes a small set of historical budget allocation episodes to form compact and informative demonstration sequences. Through few-shot prompting, it produces an initial allocation vector that indicates how the budget should be proportionally distributed across different time periods. This layer emphasizes capturing high-level strategy and distribution trends.

\textbf{Fine-grained Optimizer} focuses on refining and executing the first episode plan in a fine-grained manner. It takes as input the proposed allocation from Few Shot Reasoner and marginal ROI indicators for each time period, and aims to locally optimize the budget to maximize marginal ROI. Unlike Few Shot Reasoner, the executor is trained to be more responsive to adjustments in budget allocation. We also introduce a \textbf{sliding window mechanism} that allows Fine-grained Optimizer to dynamically adjust its strategy based on the most recent episodes' feedback records. 

\begin{algorithm}[htbp]
\caption{Dual-Phase Cooperative Budget Allocation}
\label{alg:hierarchical-rl}
\KwIn{Few-shot historical episodes $\mathcal{H}$, budget constraint $B$, periods $T$, sliding window size $w$}
\KwOut{Allocation trajectory $\mathcal{B} = \{b^{(1)}, b^{(2)}, \dots, b^{(i)}, \dots\}$, where $b^{(i)} \in \mathbb{R}^T$}

\textbf{Initialize:} $\mathcal{B} \leftarrow [\,]$, episode index $s \leftarrow 1$ \\

\tcp{Step 1: Few-shot Initialization via Few Shot Reasoner} 
Generate prompt $\mathcal{P}_1 \leftarrow$ Encode($\mathcal{H}$) \\
Initial allocation $b^{(1)} \leftarrow$ Few Shot Reasoner($\mathcal{P}_1$) \\
Append $b^{(1)}$ to $\mathcal{B}$ \\
Observe $MROI$ feedback $r^{(1)}$ \\
$s \leftarrow s + 1$ \\

\tcp{Initialize sliding window $\mathcal{W}_2$}
Extract the latest $(w-1)$ records from $\mathcal{H}$ as $\mathcal{H}_{\text{tail}}$ \\
$\mathcal{W}_2 \leftarrow \mathcal{H}_{\text{tail}} \cup \{(b^{(1)}, r^{(1)})\}$ \\

\tcp{Step 2: Fine-Grained Refinement via Fine-grained Optimizer}
\While{\textbf{True}}{
    Generate prompt $\mathcal{P}_t \leftarrow$ Encode($\mathcal{W}_{s}$) \\
    Refined allocation $b^{(s)} \leftarrow$ Fine-grained Optimizer($\mathcal{P}_t$) \\
    Append $b^{s}$ to $\mathcal{B}$ \\
    Observe marginal ROI feedback $r^{(s)}$ \\
    Update sliding window history $\mathcal{W}_{s+1}$ with $(b^{(s)}, r^{(s)})$ \\
    $s \leftarrow s + 1$ \\
    \If{termination condition met}{
        \textbf{break}
    }
}

\Return{$\mathcal{B}$}
\end{algorithm}

The overall workflow is summarized in Algorithm~\ref{alg:hierarchical-rl}. The core advantage of this two-tier architecture lies in the clear separation of concerns: the Few Shot Reasoner handles global strategic reasoning and trend estimation, while the Fine-grained Optimizer continuously refines the plan based on feedback. To further enhance LLMs' reasoning capabilities, we fine-tune these two LLMs using our proposed reinforcement learning method, enabling them to generate strong initial solutions even with limited training samples.

%% file: chaps/RLAM.tex
\section{RL Fine-tuning Strategies}
\label{sec:AERO}

\subsection{GRPO-Adaptive}
\label{sec:grpo-objective}

\subsubsection{Optimization Objective}

Given an input prompt $x$, the language model defines a policy $\pi_\theta(b \mid \mathcal{P})$ that generates an output sequence $b = (b^1, b^2, \dots, b^T)$ of the period length $T$, where the quality of the output is evaluated by a reward function $R(b)$. We define the objective as the expected discounted cumulative return under a given budget cost constraint:

\begin{equation}
\underset{\theta}{\max} \,\, J(\theta) = \mathbb{E}_{\{b^{(s)}\} \sim \pi_\theta(\cdot \mid \mathcal{P})} \left[ \sum_{s=1}^\infty \gamma^{s-1} R(b^{(s)}) \right] 
 \,\, \text{s.t.}  \,\, \forall s, \sum_{i=1}^T b^{(s)}_i = B.
\end{equation}

\noindent where $B$ is the allocation budget. The discount factor $\gamma \in (0,1]$ controls the weight of future rewards.

While KL regularization is crucial for constraining policy drift in GRPO~\cite{guo2025deepseek}, our experiments in Section~\ref{sec:kl-adaptive} reveal a key limitation: when the reference policy $\pi_{\theta_0}$ is kept static throughout training, the model’s sensitivity to structured reasoning and symbolic precision gradually deteriorates. This becomes a performance bottleneck, especially in tasks requiring rigorous multi-step reasoning.

To address this, we propose \textbf{GRPO-Adaptive (GRPO-A)}, an enhanced variant of GRPO where the reference model is periodically updated to reflect the latest policy improvements. Specifically, GRPO-A operates as follows:

\begin{itemize}
    \item After every $K$ GRPO updates, we take a snapshot of the current policy $\pi_{\theta}$,
    \item Replace the static reference $\pi_{\theta_0}$ with this updated version,
    \item Reset the KL regularization to reflect divergence from the new baseline.
\end{itemize}

The resulting optimization objective becomes:

\begin{equation}
\underset{\theta}{\max} \,\, J_{\text{GRPO-A}}(\theta) = J(\theta) - \beta \, \mathbb{E}_{\mathcal{P}} \left[ \mathbb{D}_{\mathrm{KL}} \left( \pi_\theta(\cdot \mid \mathcal{P}) \,\| \underbrace{\pi_{\theta_{\text{ref}}}(\cdot \mid \mathcal{P})}_{\text{periodically updated}} \right) \right],
\end{equation}

\noindent where $\beta$ controls the trade-off between reward maximization and policy stability, and $\mathbb{D}_{\mathrm{KL}}[\cdot \| \cdot]$ is the KL divergence.

This dynamic adjustment allows the model to continue learning from its improved reasoning abilities while still benefiting from KL stabilization. It prevents over-regularization by outdated references and promotes a more adaptive, stable optimization process in later-stage fine-tuning.

\subsubsection{Optimization loss}

To practically optimize the policy, we adopt a clipped advantage-weighted objective as used in GRPO. For a batch of $G$ trajectories, the loss is:

\begin{align}
\mathcal{L}_{\text{GRPO-A}}(\theta) = 
& -\mathcal{L}_{\text{adv}}(\theta) 
+ \beta\, \mathbb{D}_{\text{KL}} \left[ \pi_{\theta} \,\|\, \pi_{\text{ref}} \right] \label{eq:grpo_total} \\
\quad 
\mathcal{L}_{\text{adv}}(\theta) = 
& \frac{1}{G} \sum_{i=1}^{G} \frac{1}{|o_i|} 
\sum_{t=1}^{|o_i|} \min \Bigg(
\frac{\pi_{\theta}(o_{i,t} \mid \mathcal{P})}{\pi_{\theta_{\text{old}}}(o_{i,t} \mid \mathcal{P})} \hat{A}_{i,t}, \nonumber \\
&\quad\quad\quad\quad
\text{clip} \left( 
\frac{\pi_{\theta}(o_{i,t} \mid \mathcal{P})}{\pi_{\theta_{\text{old}}}(o_{i,t} \mid \mathcal{P})}, 
1 - \epsilon, 1 + \epsilon 
\right) \hat{A}_{i,t}
\Bigg) \label{eq:grpo_adv}
\end{align}

\begin{equation}
\label{eqa:Loss}
\hat{A}_i = \frac{r_i - \mu_r}{\sigma_r},
\quad \text{with} \quad
\mu_r = \frac{1}{G} \sum_{j=1}^{G} r_j, \,\,\,
\sigma_r = \sqrt{ \frac{1}{G} \sum_{j=1}^{G} (r_j - \mu_r)^2 }.
\end{equation}

Here, $\pi_{\theta_{\text{old}}}$ is the previous policy snapshot, $\pi_{\text{ref}}$ is the reference policy updated periodically, and $\epsilon$ is the clipping parameter.

\subsection{Fine-tuning Procedure with GRPO-Adaptive}

To simulate budget allocation in a realistic yet diverse training setting, we design a controlled multi-environment sampling procedure. Within each environment instance, we simulate a single episode budget allocation process. After a fixed number of sampling steps, the model switches to a new environment to emulate a different budget planning scenario. This avoids contamination between training and evaluation phases and encourages the model to generalize across settings rather than memorize a fixed dataset.

Every fixed $S$ steps, we randomly resample or procedurally generate a new environment, which includes $T$ novel marginal ROI curves. The objective is to ensure that the model learns a generalizable ability to identify better solutions, rather than overfitting to any single distribution. Once the model exhibits stable performance in simulation, it can be deployed directly in real environments. % to perform inference without requiring further simulation-based rollouts.

In Equation \ref{eq:grpo_adv}, for each prompt, we sample $G$ candidate outputs in parallel within each environment. The “preferred” output is selected from feasible completions with higher reward. Using normalized group-wise advantage estimation and a clipped PPO-style objective, we update the policy to increase the relative probability of high-quality completions while suppressing low-reward outputs. After each round, the best-performing candidate is added to the trajectory history and the sampling-update cycle is repeated until $S$ total steps are completed. The overall training process is formally described in Algorithm~\ref{alg:grpo} in Appendix~\ref{appendix:GRPO}.

\subsection{Policy Interface and Reward Design}
\label{sec:policy-reward}

% While our task setup does not conform to a standard Markov Decision Process (MDP), reinforcement learning remains effective in optimizing language model behaviors via preference-driven feedback.

In our framework, two LLMs are coordinated: the first LLM provides an initial allocation strategy on the first day, while the second LLM refines and outputs updated allocation vectors from the second day onward. Despite differing prompt formats, both models share a unified action interface. This section introduces the formal design of input state representations, output actions, and reward functions tailored to budget allocation.

\subsubsection{State and action format}

The state is defined as the input prompt $\mathcal{P}$, which includes the task description as well as the sequence of optimal allocations selected in previous steps. The action is defined as a budget allocation vector of length $N$:

\begin{equation}
b = [b_1,\, b_2,\, \dots,\, b_T],
\end{equation}
where $b_i$ denotes the budget assigned to the $i$-th period.

To extract the action from the LLM output, we first identify the content enclosed in \texttt{\textless answer\textgreater}~\ldots~\texttt{\textless /answer\textgreater}, and then use a regular expression to extract the vector pattern \texttt{[}$b_1$, $b_2$, \ldots, $b_T$\texttt{]}.

\subsubsection{Reward Function}

A major challenge in LLM-based budget allocation lies in the numerical sensitivity and structural complexity of the task. Small deviations in token-level outputs can lead to substantial losses in overall performance, and the underlying reward landscape is often non-convex and highly structured. To mitigate these issues, we design a composite reward function that incorporates both structural decomposition and auxiliary shaping mechanisms to enhance learning stability and policy robustness.

For each action $b \in \mathbb{R}^T$, the core reward signal $R(b)$ is composed of three terms. The first term, the environment reward \( R_{\mathrm{env}} \), encourages consistent marginal returns across time segments. Specifically, let the marginal reward at time segment \( i \) be \( r_i = MROI_i(p_i) \), and let the average marginal reward be \( \bar{r} = \frac{1}{N} \sum_{i=1}^N r_i \). We define the environment reward as:
\begin{equation}
R_{\mathrm{env}}(b) = -\alpha \sum_{i=1}^N \left| r_i - \bar{r} \right|,
\end{equation}
where \( \alpha \) is a scaling coefficient controlling the model’s sensitivity to reward variance.

To ensure the validity of generated actions, we define a constraint penalty \( R_{\mathrm{constraint}} \) that enforces two conditions: the budget vector must have the correct dimensionality and the total allocation should be close to the total budget $B$. This penalty is defined as:
\begin{equation}
R_{\mathrm{constraint}}(b) = 
\begin{cases}
-M, & \text{if } \dim(b) \neq N, \\
-\left| \sum_{i=1}^N p_i - B \right|, & \text{otherwise}.
\end{cases}
\end{equation}

In addition, we introduce a bonus reward term \( R_{\mathrm{bonus}} \) to encourage the model to refine its allocation strategy based on last historical performance adaptively. Let \( c \in \mathbb{R}^N \) be the last allocations. We define two subsets of time segments: \( \mathcal{H}_+ \) for those with high historical rewards, and \( \mathcal{H}_- \) for those with low historical rewards. Given a minimum adjustment threshold \( \delta > 0 \) and a clipping bound \( \tau > 0 \), the bonus reward for each segment is computed as follows. For each \( b^i \in \mathcal{H}_+ \), if the current allocation \( b^i > c^i + \delta \), we assign a positive adjustment reward:
\[
R_{\text{bonus}, i}^{+} = \min\left( |b^i - c^i|, \tau \right).
\]
For each \( i \in \mathcal{H}_- \), if \( b^i < c^i - \delta \) and \( b^i > 0 \), we also assign a complementary bonus:
\[
R_{\text{bonus}, i}^{-} = \min\left( |b^i - c^i|, \tau \right).
\]
The total bonus reward is given by:
\[
R_{\text{bonus}}(b) = \sum_{b^i \in \mathcal{H}_+} R_{\text{bonus}, b^i}^{+} + \sum_{b^i \in \mathcal{H}_-} R_{\text{bonus}, b^i}^{-}.
\]

Finally, the full scalar reward used to guide policy updates is defined as the sum of all components:
\begin{equation}
R(b) = R_{\mathrm{env}}(b) + R_{\mathrm{constraint}}(b) + R_{\mathrm{bonus}}(b).
\end{equation}

%% file: chaps/evaluation.tex
\section{Evaluation}
\label{sec:evaluation}

\subsection{Experimental Setup}

This study evaluates the model in two types of environments.\footnote{Our implementation is available at \url{https://github.com/mx-song/DARA}} The first is the \textbf{real data environment}, where we use online advertising data from a leading global e-commerce platform to construct a highly realistic simulation of the advertisers. This environment reflects the dynamic variations in the ad allocation process, ensuring that the experimental results have strong practical relevance. The second type is the \textbf{synthetic data environment}. We design a series of $MROI$ functions with diminishing marginal returns to test the model's generalization ability in a controlled setting. These two environments are designed to assess the model's performance and robustness from different perspectives. Each episode corresponds to a single day’s budget allocation across multiple time periods in real-world ad planning.

All training tasks were conducted on enterprise-level private servers with hardware configuration of 8 NVIDIA H20 GPUs (each with 96GB of memory). The baselines and main hyperparameters used in the experiments are listed in the Appendix \ref{apen:Hyperparameters}. To ensure the reliability of the results, each experiment was repeated five times, and the mean performance along with the 95\% confidence interval is reported. The shaded areas in the plots represent these confidence intervals. The \textbf{variance of MROI} measures the balance of allocation across time periods and reflects the stability and uniformity of the budget distribution. 

\subsection{Result Analysis}

\begin{figure}
    \centering
    % \vspace{-12pt}
    \includegraphics[width=0.35\textwidth]{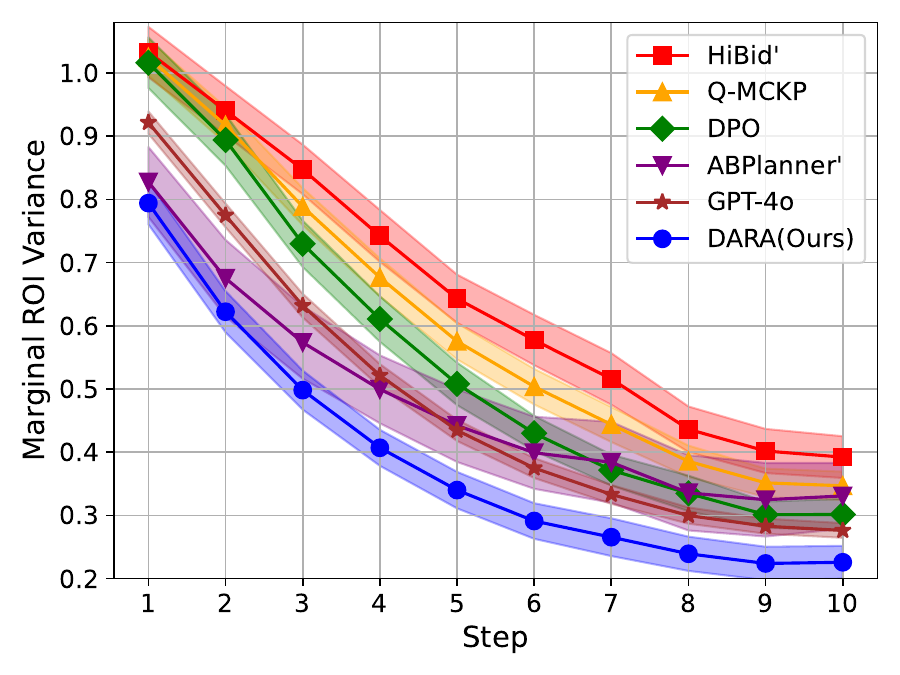}
    \captionsetup{aboveskip=2pt, belowskip=-5pt}
    \caption{Performance of different algorithms in reducing marginal ROI variance.}
    \label{fig:algorithm-result}
\end{figure}

After training on the simulation environment, we evaluate the performance of our method in the real-world data environment. As shown in Figure~\ref{fig:algorithm-result}, our method significantly outperforms all baseline algorithms across all steps in terms of reducing the marginal ROI variance. This demonstrates the strong stability and adaptability of our approach in dynamic budget allocation environments.

Compared to DPO, which rely on either supervised or preference-aligned fine-tuning, our method exhibits consistently lower variance, indicating a more balanced and effective distribution across time periods. Notably, while DPO improves generalization via token-level preference alignment, this approach is fundamentally limited by its static modeling structure and lack of dynamic feedback responsiveness. Moreover, DARA demonstrates particularly significant improvements in reducing marginal ROI variance during the later stages of allocation. This advantage arises from its fine-grained optimization phase, where recent allocation outcomes provide more reliable guidance than initial few-shot exemplars. By explicitly separating early generalization from late-stage adaptation, our dual-phase design enables the model to focus on precise, context-aware refinements—an ability that single-stage baselines inherently lack.

Compared with ABPlanner’, which also uses RL for budget strategy design, our method achieves better variance reduction, especially in later steps. This improvement is attributed to our method’s enhanced numerical sensitivity and the structured separation, which allows each LLM to specialize and avoid underfitting or overgeneralization.

\subsection{Ablation Performance}
\label{sec:ablation}

\begin{figure}
    \centering
    % \vspace{-12pt}
    \includegraphics[width=0.35\textwidth]{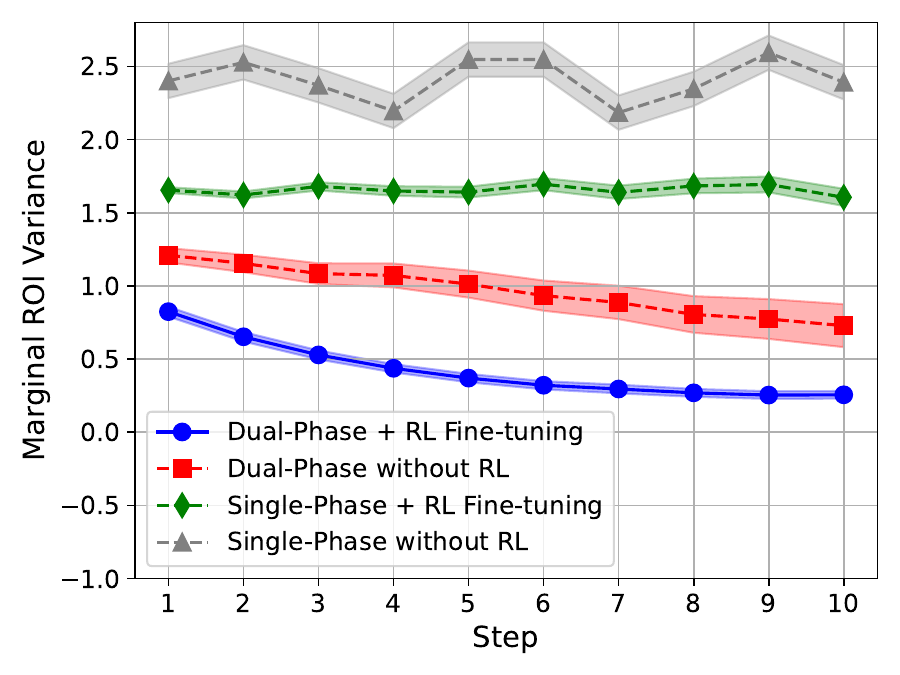}
    \captionsetup{aboveskip=2pt, belowskip=-5pt}
    \caption{Impact of dual-phase architecture and RL fine-tuning on marginal ROI variance.}
    \label{fig:ablation-result}
\end{figure}

To investigate the effectiveness of each component in our framework, we conduct a comprehensive ablation study with four different configurations. The corresponding results are shown in Figure~\ref{fig:ablation-result}.

We observe that directly applying a single LLM to perform end-to-end budget planning yields the poorest performance, exhibiting the highest marginal ROI variance throughout the entire allocation time horizon. This clearly reflects the inherent limitations of large language models in handling numerically sensitive and temporally dynamic tasks. Although LLMs demonstrate strong few-shot reasoning capabilities, they tend to generalize poorly in complex allocation scenarios where subtle reward shifts need to be captured and responded to accurately.

Introducing RL fine-tuning into the single-phase setup offers a slight improvement, confirming that reinforcement learning can enhance the numerical reasoning ability of the model to some extent. However, the gain remains limited. This is because the single LLM is still responsible for both global few-shot reasoning and local decision optimization, leading to interference and capacity bottlenecks when faced with long-horizon, structured decisions.

In contrast, when we adopt the dual-phase architecture without RL, the performance improves substantially and consistently. By explicitly separating the reasoning process into two specialized LLM agents—one focusing on early-stage few-shot generalization and the other on late-stage fine-grained adaptation—the model is better able to handle the heterogeneous requirements of the task. This confirms our hypothesis that task decomposition is critical for complex budget allocation.

Finally, our full model, which incorporates RL fine-tuning into both LLMs, achieves the best overall performance. The Few Shot Reasoner benefits from RL by learning to generate more strategic and generalizable initial allocation plans, while the Fine-grained Optimizer becomes more numerically sensitive and responsive to feedback, thanks to the reinforcement signal. The collaboration between these two specialized agents allows the model to learn both abstract patterns and local refinements, resulting in a significantly lower variance and more stable performance. The performance boost brought by RL is significantly amplified when applied within the dual-phase architecture. While RL alone offers modest improvements in the single-phase setup, its effect becomes substantially more pronounced when each LLM is structurally disentangled and specialized. This observation highlights that its presence does not solely determine the effectiveness of RL in budget planning tasks, but also by how it is integrated—underscoring the importance of tailored architectural design for fully leveraging the benefits of policy learning.

\subsection{Sensitivity Analysis}

\label{sec:sensitivity}

\subsubsection{Time periods}

\begin{figure}
    \centering
    % \vspace{-12pt}
    \includegraphics[width=0.4\textwidth]{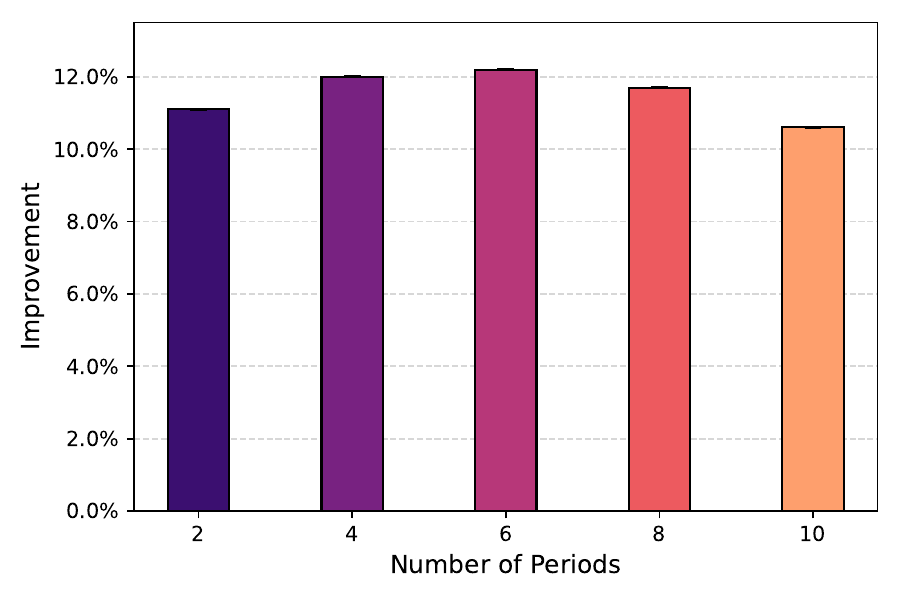}
    % \captionsetup{aboveskip=2pt, belowskip=-15pt}
    \caption{Experimental results with different numbers of periods.}
    \label{fig:cross}
\end{figure}

To assess the robustness of our approach to the granularity of temporal partitioning, we conduct a sensitivity analysis by varying the number of time periods into which the total budget horizon is divided. Specifically, we evaluate the performance of our model under 5 different configurations: 2, 4, 6, 8, and 10 time periods. The results, presented in Figure~\ref{fig:cross}, report the percentage improvement of our method over the strongest baseline (ABPlanner’) in each setting.

We observe that our model consistently achieves performance improvements across all configurations, ranging from approximately 10.6\% to 12.2\%. The performance curve exhibits only mild fluctuations as the number of periods increases, indicating that our method is not sensitive to the specific temporal granularity chosen. In particular, the improvement is most pronounced when the number of periods is 6, aligning with typical real-world ad scheduling practices where campaigns are often divided into early, middle, and late phases. This also demonstrates the flexibility of our framework in capturing temporal dependencies at multiple resolutions.

\subsubsection{Effect of KL Reference Refresh Strategy}

\label{sec:kl-adaptive}

\begin{figure}
    \centering
    % \vspace{-12pt}
    \includegraphics[width=0.4\textwidth]{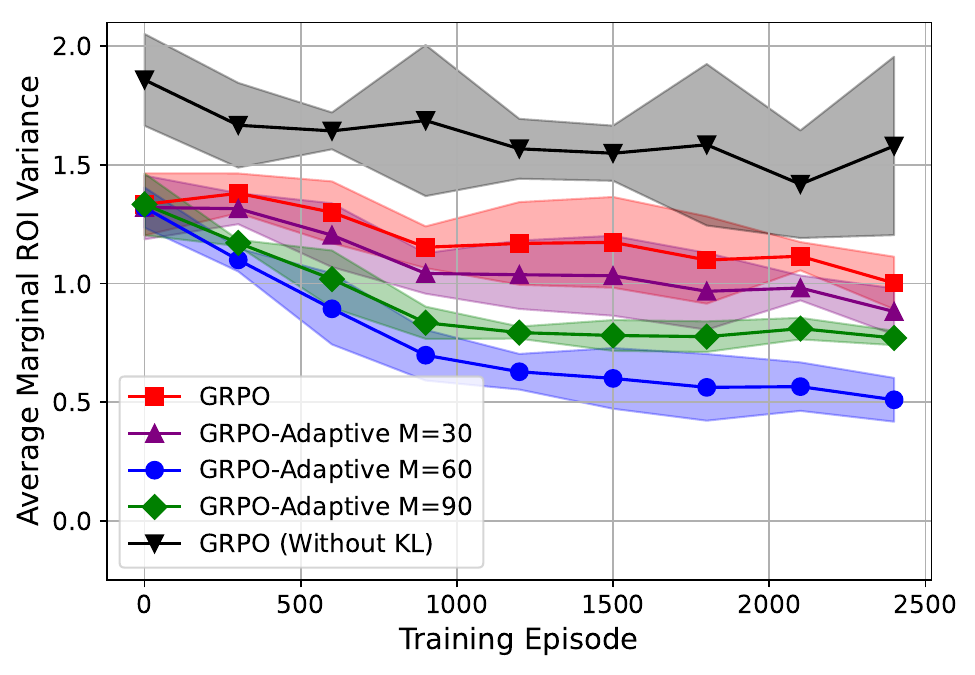}
    % \captionsetup{aboveskip=2pt, belowskip=-10pt}
    \caption{Experimental results with different frequencies of updating the reference model.}
    \label{fig:grpo-kl}
\end{figure}

To analyze the effect of KL regularization in GRPO-based fine-tuning, we compare standard GRPO with several GRPO-Adaptive variants under different KL reference update frequencies, including updates every $M=30$, $60$, and $90$ iterations, as well as a variant that entirely removes the KL term. As shown in Figure~\ref{fig:grpo-kl}, our full model (GRPO-Adaptive, $M=60$) achieves the lowest Marginal ROI variance across training episodes, indicating that periodically refreshing the reference model substantially improves training stability and alignment. This design allows the target policy to continuously benefit from improved generations while maintaining a sufficiently stable regularization anchor, striking an effective balance between adaptability and constraint.

When the reference is updated too frequently (e.g., $M=30$), the gain over vanilla GRPO becomes marginal, which we attribute to instability during early training: frequent updates tend to anchor the KL term to partially optimized or noisy generations, including incoherent reasoning and erratic numerical decisions. In contrast, updating the reference too slowly (e.g., $M=90$) delays alignment with the improved policy and slows convergence, although it still outperforms static GRPO. Removing the KL term altogether leads to significantly higher variance and unstable training, demonstrating that KL regularization is essential for constraining the policy within a meaningful distributional bound rather than serving as a mere stabilizer. Overall, these results confirm that an intermediate update frequency provides the most effective trade-off, and that GRPO-Adaptive offers a simple yet robust improvement over fixed-reference regularization schemes.

%% file: chaps/appendix/SafetyAnalysis.tex
\appendix
\section{Proof of Marginal ROI Optimality Condition}
\label{appendix:A}

In Section~2, we state that under concave return functions, the optimal allocation under a fixed total budget satisfies that all marginal ROIs (i.e., first derivatives of return functions) are equal. We provide a formal proof of this result below.

\paragraph{Setting.} Let $B$ denote the total available budget to be allocated across $T$ time periods. For each period $t \in \{1, \dots, T\}$, let $b_t$ denote the allocated budget, and let $v_t(b_t)$ denote the return function in that period. The optimization objective is:
\begin{equation}
\max_{\{b_t\}} \sum_{t=1}^T v_t(b_t) \quad \text{subject to} \quad \sum_{t=1}^T b_t = B, \quad b_t \geq 0
\end{equation}

\paragraph{Assumptions.} For each $t$, the return function $v_t(b)$ is:
\begin{itemize}
    \item Differentiable,
    \item Strictly increasing: $v_t'(b) > 0$,
    \item Concave: $v_t''(b) < 0$.
\end{itemize}

\paragraph{Claim.} At the optimum $\{b_t^*\}$, the marginal ROI must be equal across all periods:
\begin{equation}
v_1'(b_1^*) = v_2'(b_2^*) = \cdots = v_T'(b_T^*)
\end{equation}

\paragraph{Proof.}  
Assume, for contradiction, that there exists an optimal allocation $\{b_t^*\}$ and two indices $i \neq j$ such that:
\begin{equation}
v_i'(b_i^*) > v_j'(b_j^*)
\end{equation}

We construct a new allocation $\{\tilde{b}_t\}$ as:
\begin{equation}
\tilde{b}_i = b_i^* + \delta, \quad \tilde{b}_j = b_j^* - \delta, \quad \tilde{b}_t = b_t^* \quad \text{for } t \notin \{i,j\}
\end{equation}
where $\delta > 0$ is sufficiently small so that $\tilde{b}_i, \tilde{b}_j \geq 0$.

Note that $\sum_{t=1}^T \tilde{b}_t = \sum_{t=1}^T b_t^* = B$, so the new allocation remains feasible.

The change in total return is:
\begin{equation}
\Delta R = \left[ v_i(b_i^* + \delta) - v_i(b_i^*) \right] + \left[ v_j(b_j^* - \delta) - v_j(b_j^*) \right]
\end{equation}

Using Taylor expansion (and concavity) around $b_i^*$ and $b_j^*$:
\begin{equation}
v_i(b_i^* + \delta) - v_i(b_i^*) = v_i'(b_i^*) \delta - \frac{1}{2} |v_i''(\xi_i)| \delta^2
\end{equation}
\begin{equation}
v_j(b_j^* - \delta) - v_j(b_j^*) = -v_j'(b_j^*) \delta - \frac{1}{2} |v_j''(\xi_j)| \delta^2
\end{equation}
for some $\xi_i \in (b_i^*, b_i^* + \delta)$ and $\xi_j \in (b_j^* - \delta, b_j^*)$.

Thus,
\begin{equation}
\Delta R = \left[v_i'(b_i^*) - v_j'(b_j^*)\right] \delta - \frac{1}{2} \left(|v_i''(\xi_i)| + |v_j''(\xi_j)|\right) \delta^2
\end{equation}

Since $v_i'(b_i^*) > v_j'(b_j^*)$, the first-order term dominates the second-order term for small enough $\delta$, so $\Delta R > 0$. Hence, $\{\tilde{b}_t\}$ yields strictly higher total return, contradicting optimality.

Therefore, the optimal solution must satisfy:
\begin{equation}
v_1'(b_1^*) = v_2'(b_2^*) = \cdots = v_T'(b_T^*)
\end{equation}

\section{Prompt Examples}
\label{apen:PromptExamples}

We present two representative prompt templates used for guiding the language model's decision-making process.

\subsection*{Prompt A: Few Shot Reasoner}
\begin{lstlisting}
You are given a total budget of {TOTAL} to allocate across {NUM} time periods. Based on your last attempt, identify the time period with the **lowest reward**, and reallocate some of its budget to the time period with the **highest reward**. Keep the allocations for other periods unchanged.

Last attempt: {few-shot data}

Please output the new allocation:
<reason>
...
</reason>
<answer>
[y1, y2, ..., y{NUM}]
</answer>

Your Response:
<reason>
\end{lstlisting}

\subsection*{Prompt B: Fine-grained Optimizer}
\begin{lstlisting}
You are given a budget allocation task across {NUM} time periods. The total budget is {TOTAL}, and it must be fully used - no more, no less. In every time period, allocating more budget results in smaller reward. Your goal is to equalize the reward across all periods.

You are provided with the last some rounds of allocations and observed rewards. Use this historical data to identify:
- Which periods likely has lower reward, reduce allocation slightly.
- Which periods likely has higher reward, increase allocation slightly.

Do not resort to a naive and robust uniform allocation. Instead, carefully analyze the underlying patterns and allocate the budget based on the relationship between historical data and observed rewards.

Last attempt: {sliding-window data}

Please output the new allocation:
<reason>
...
</reason>
<answer>
[y1, y2, ..., y{NUM}]
</answer>

Your Response:
<reason>
\end{lstlisting}

\newpage
\section{Training Procedure of GRPO-Adaptive}
\label{appendix:GRPO}
\FloatBarrier
\begin{algorithm}[h]
\caption{GRPO-Adaptive in DARA}
\label{alg:grpo}
\KwIn{Pretrained policy $\pi_{\theta_0}$, prompt distribution $\mathcal{P}(Q)$}
\KwOut{Updated policy $\pi_\theta$}

Initialize policy parameters $\theta \leftarrow \theta_0$ \\
Initialize KL reference baseline $\pi_{\text{ref}} \leftarrow \pi_{\theta_0}$ \\
Set hyperparameters: group size $G$, clip range $\epsilon$, KL weight $\beta$, baseline update period $M$ \\

\For{iteration $= 1$ to $N$}{
  \tcp{Step 1: Sample batch and generate outputs}
  Sample prompts $\{q_1, \dots, q_B\} \sim \mathcal{P}(Q)$ \\
  For each $q$, generate outputs $\{o_i^q\}_{i=1}^G \sim \pi_{\theta}(\cdot \mid q)$

  \tcp{Step 2: Compute reward for each output}
  $r_{i} \leftarrow R(q, o_{i}^q)$

  \tcp{Step 3: Compute normalized advantage}
  $\hat{A}_{i,t} \leftarrow \frac{r_{i,t} - \mu}{\sigma}$ where $\mu = \text{mean}(\{r_{i,t}\})$, $\sigma = \text{std}(\{r_{i,t}\})$

\tcp{Step 4: Compute loss}
 $\mathcal{L}_{\text{GRPO-A}}(\theta) = 
 - L_{Adv}(\theta) + \beta \cdot D_{\text{KL}}(\pi_\theta \| \pi_{\text{ref}})$

  \tcp{Step 5: Update policy parameters}
  $\theta \leftarrow \theta - \alpha \cdot \nabla_\theta \left( \frac{1}{B} \sum_q \mathcal{L}_{\text{GRPO-A}}(q) \right)$

  \tcp{Step 6: Refresh KL baseline}
  \If{iteration mod $M$ == 0}{
    $\pi_{\text{ref}} \leftarrow \texttt{deepcopy}(\pi_\theta)$
  }
}
\end{algorithm}

\FloatBarrier
\section{Hyperparameter and Baseline Configuration}
\label{apen:Hyperparameters}

The hyperparameter settings used in DARA are summarized in Table \ref{tab:hyperparameters}. To comprehensively assess the performance of our proposed method, we have selected five algorithms for comparison:

\begin{itemize}
    \item \textbf{ABPlanner'}~\cite{duan2025adaptable}: ABPlanner' is a simplified version of ABPlanner, designed to control for variables by focusing solely on the reinforcement learning component and excluding the controller design. ABPlanner' effectively demonstrates the strategy optimization capability of reinforcement learning algorithms in budget allocation tasks.
    
    \item \textbf{HiBid'}~\cite{wang2023hibid}: HiBid' is a simplified version of HiBid (Wang et al., 2023), which uses a hierarchical reinforcement learning algorithm to learn both high-level budget planning and low-level bidding strategies. Unlike the original HiBid, HiBid' focuses only on the high-level budget planning component, making it suitable for improving fixed underlying auto-bidder systems.
    
    \item \textbf{Q-MCKP}~\cite{li2018efficient}: Q-MCKP  is an algorithm that discretizes the budget space into multiple bins and independently learns the Q-value function for each stage. It solves the optimal budget allocation based on the learned Q-values.
    
    \item \textbf{DPO Fine-tuning}~\cite{rafailov2023direct}: Direct Preference Optimization (DPO) fine-tuning is a preference-alignment-based fine-tuning method that enhances the model's generalization ability in few-shot tasks by extracting key tokens from preference and rejection pairs during training.
    
    \item \textbf{GPT-4o}~\cite{hurst2024gpt}: GPT-4o is one of the most advanced LLMs developed by OpenAI. It demonstrates strong in-context learning capabilities across a wide range of tasks. In this work, GPT-4o performs reasoning and decision-making solely based on prompt engineering, without any parameter updates.
\end{itemize}

\begin{table}[H]
    \centering
    \caption{Hyperparameters and their values}
    \begin{tabular}{|l|l|}
        \hline
        \multicolumn{1}{|c|}{\textbf{Hyperparameter}} & \multicolumn{1}{c|}{\textbf{Value}} \\ \hline
        \multicolumn{1}{|c|}{Number of training iterations, $N$} & \multicolumn{1}{c|}{500} \\ \hline
        \multicolumn{1}{|c|}{Number of baseline update period, $M$} & \multicolumn{1}{c|}{60} \\ \hline
        \multicolumn{1}{|c|}{Number of time periods, $T$} & \multicolumn{1}{c|}{6} \\ \hline
        \multicolumn{1}{|c|}{Batch size, $batch\_size$} & \multicolumn{1}{c|}{8} \\ \hline
        \multicolumn{1}{|c|}{Number of generations, $num\_generations$} & \multicolumn{1}{c|}{3} \\ \hline
        \multicolumn{1}{|c|}{Max completion length, $max\_completion\_length$} & \multicolumn{1}{c|}{500} \\ \hline
        \multicolumn{1}{|c|}{KL regularization coefficient, $\beta$} & \multicolumn{1}{c|}{0.04} \\ \hline
        \multicolumn{1}{|c|}{Learning rate, $lr$} & \multicolumn{1}{c|}{$5 \times 10^{-6}$} \\ \hline
        \multicolumn{1}{|c|}{Scaling factor for preference loss, $\mu$} & \multicolumn{1}{c|}{1} \\ \hline
        \multicolumn{1}{|c|}{Epsilon for PPO clipping, $\epsilon$} & \multicolumn{1}{c|}{0.1} \\ \hline
        \multicolumn{1}{|c|}{Minimum Adjustment threshold, $\delta$} & \multicolumn{1}{c|}{0.2} \\ \hline
        \multicolumn{1}{|c|}{Clipping bound, $\tau$} & \multicolumn{1}{c|}{0.2} \\ \hline
        \multicolumn{1}{|c|}{Scaling coefficient, $\tau$} & \multicolumn{1}{c|}{0.5} \\ \hline
        \multicolumn{1}{|c|}{Total budget, $B$} & \multicolumn{1}{c|}{6} \\ \hline
        \multicolumn{1}{|c|}{Number of tries per environment, $N\_TRY$} & \multicolumn{1}{c|}{10} \\ \hline
    \end{tabular}
    \label{tab:hyperparameters}
\end{table}

\section{Response Sample}
\label{apen:ResponseSample}

Below is a representative example of the model's generated reasoning and corresponding action in a decision-making round:

\begin{lstlisting}
<reason>
Looking at the historical data, I notice that periods 3 and 5 have consistently low rewards (e.g., 0.0000 in multiple rounds), indicating that allocating more budget to these periods would result in smaller rewards. Therefore, I should reduce the allocation for these periods.
Conversely, periods 4 and 6 have higher rewards (e.g., 0.2469, 0.2693, 0.4895 in previous rounds), suggesting that increasing the budget for these periods may yield higher rewards. Thus, I should increase the allocation for these periods.
Additionally, period 2 has shown some reward (0.0733, 0.0726), so I should slightly increase its allocation to take advantage of the modest reward.
However, I must ensure the total allocation sums to 6. After adjusting allocations based on the above observations, I need to balance the budget carefully to maintain the total allocation of 6.
</reason>
<answer>
[0.93, 1.18, 0.77, 1.19, 0.75, 1.18]
</answer>
\end{lstlisting}